# The Virtual Electromagnetic Interaction between Digital Images for Image Matching with Shifting Transformation

XIAODONG ZHUANG[1,2] and N. E. MASTORAKIS[1,3,4]
1. WSEAS Research Department, Agiou Ioannou Theologou 17-23, 15773, Zografou,
Athens, GREECE (xzhuang@worldses.org, http://research-xzh.cwsurf.de/)
2. Automation Engineering College, Qingdao University, Qingdao, CHINA
3. Department of Computer Science, Military Institutions of University Education,
Hellenic Naval Academy, Terma Hatzikyriakou, 18539, Piraeus, GREECE
mastor@wseas.org    http://www.wseas.org/mastorakis
4. Technical University of Sofia, BULGARIA

*Abstract:* -A novel way of matching two images with shifting transformation is studied. The approach is based on the presentation of the virtual edge current in images, and also the study of virtual electromagnetic interaction between two related images inspired by electromagnetism. The edge current in images is proposed as a discrete simulation of the physical current, which is based on the significant edge line extracted by Canny-like edge detection. Then the virtual interaction of the edge currents between related images is studied by imitating the electro-magnetic interaction between current-carrying wires. Based on the virtual interaction force between two related images, a novel method is presented and applied in image matching for shifting transformation. The preliminary experimental results indicate the effectiveness of the proposed method.

## 1 Introduction

Image registration is one of the research focuses in image processing, which is applied in many practical tasks such as remote sensing, medical image, robot vision, photography, etc [1-9]. The registration can facilitate the integration of useful data in two related but separated images, so that the complementary information in the images can be fully exploited in practical tasks. The relative position of the two images can also be estimated by image matching. Current image registration methods can be divided into several categories from different viewpoints. Some methods are based on grayscale or color matching, while others are based on feature extraction and matching [1-9]. The nature of transformation from one image to another may be one of the following cases: rigid, affine, projective or curved [1-9]. Different methods have their own advantages and disadvantages respectively. Most methods are based on the searching for similarities between the matched parts in the two images. Currently, it is relatively difficult to implement the registration both very rapidly and very accurately (on commonly-used PCs). Moreover, largely mismatched images may invalidate many registration methods. On the other hand, most methods are not full-automatic because they need manual intervention such as setting and adjusting program parameters.

Matching is the core issue in image registration. Traditional registration methods are based on the similarity between the matched areas in two images. Quite different from previous matching methods, this paper introduces a novel idea for matching - the virtual force which reflects the transformation between two images. In the physical world, there exist various interaction forces. Some physical interaction mechanisms can be exploited in image processing tasks if properly configured and simulated. The electro-magnetic interaction is one of them. In recent years, the electro-magnetics inspired methods have become a new branch in novel image processing techniques [10-17]. In this paper, a novel method is studied for image matching inspired by the electro-magnetic interaction. In physics, a current-carrying wire can produce its magnetic field, which can further apply magnetic force on another current-carrying wire. Inspired by the physical phenomena of the attraction between two current-carrying wires, an automatic image matching method is proposed for shifting transformation. In the method, one image produces a virtual magnetic field and puts virtual magnetic force on the other image. The force can represent the relative transformation between the two images, and the matching can be implemented by following the guidance of the virtual force.

## 2 The Physical Interaction between Current-carrying Wires

### 2.1 The magnetic field

In physics, the description of the magneto-static field is given by the Biot-Savart law [18,19], where the source of the magnetic field is the current of arbitrary shapes which is composed of current elements. A current element $I\vec{dl}$ is a vector representing a very small part of the whole current, whose magnitude is the arithmetic product of $I$ and $dl$ (the length of a small section of the wire). The current element has the same direction as the current flow on the wire. The magnetic field (i.e. the magnetic induction) generated by a current element $I\vec{dl}$ is as following [18,19]:

$$\vec{dB} = \frac{\mu_0}{4\pi} \cdot \frac{I\vec{dl} \times \vec{r}}{r^3} \quad (1)$$

where $\vec{dB}$ is the magnetic induction vector at a space point. $I\vec{dl}$ is the current element. $r$ is the distance between the space point and the current element. $\vec{r}$ is the vector from the current element to the space point. The operator $\times$ represents the cross product. The direction of the magnetic field follows the right-hand rule [18,19]. The direction distribution of the magnetic field on the 2D plane where the current element lies is shown in Fig. 1.

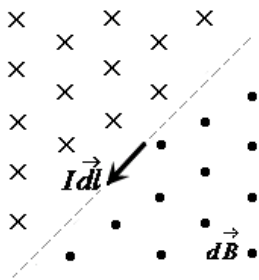

**Fig. 1 The magnetic field's direction of a current element on the 2D plane**

The magnetic field generated by the current in a wire of arbitrary shape is the accumulation of the fields generated by all the current elements on the wire [18,19]:

$$\vec{B} = \int_D \vec{dB} = \int_D \frac{\mu_0}{4\pi} \cdot \frac{I\vec{dl} \times \vec{r}}{r^3} \quad (2)$$

where $\vec{B}$ is the magnetic induction vector at a space point generated by the whole current of the wire. $D$ is the area where the current element exists. $\vec{dB}$ is the magnetic field generated by each current element in $D$.

### 2.2 The force on current-carrying wire in stable magnetic field

In electro-magnetic theory, the magnetic field applies force on moving charges, which is described by the Lorentz force. Derived from the Lorentz force, the magnetic force on a current element is as following [18,19]:

$$d\vec{F} = I\vec{dl} \times \vec{B} \quad (3)$$

where $Idl$ is the current element and $dF$ is the magnetic force on it caused by the magnetic induction $B$. The direction of $dF$ satisfies the "left hand rule", which is shown in Fig. 2. In Fig. 2, the vector of $dF$ is perpendicular to both $B$ and $Idl$.

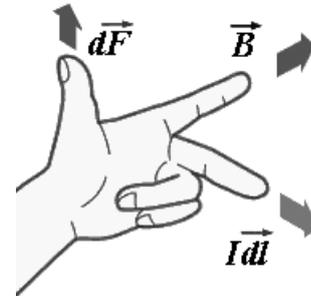

**Fig. 2 The directions of $B$, $Idl$ and $dF$**

A current-carrying wire of arbitrary shape consists of many current elements. The magnetic force on a wire is the summation of the force $dF$ on each of its current elements, which is as following [18,19]:

$$\vec{F} = \int_C I\vec{dl} \times \vec{B} \quad (4)$$

where $C$ is the integration path along the wire, and $F$ is the total magnetic force on the wire.

### 2.3 The electromagnetic interaction between two current-carrying wires

The current-carrying wire can generate magnetic field, meanwhile the magnetic field can put force on another wire. Thus there is interaction force between two current-carrying wires. In electro-magnetic theory, the Ampere force between two wires of arbitrary shapes is based on the line integration and combines the Biot-Savart law and Lorentz force in one equation as following [18,19]:

$$\vec{F}_{12} = \frac{\mu_0}{4\pi} \int_{C_1} \int_{C_2} \frac{I_1 d\vec{l}_1 \times (I_2 d\vec{l}_2 \times \vec{r}_{21})}{r_{21}^3} \quad (5)$$

where $F_{12}$ is the total force on wire1 due to wire2. $\mu_0$ is the magnetic constant. $I_1 dl_1$ and $I_2 dl_2$ are the current elements on wire1 and wire 2 respectively. $r_{21}$ is the vector from $I_2 dl_2$ to $I_1 dl_1$. $C_1$ and $C_2$ are the integration paths along the two wires respectively.

Equation (5) is virtually the accumulation of the magnetic force on the current elements in wire1 put by the current elements in wire2. If such interaction is simulated on computers, the continuous wires should be discretized, and the integration in Equation (5) should be discretized to summation:

$$\vec{F}_d = \frac{\mu_0}{4\pi} \sum_{\vec{T}_{1j} \in C_1} \sum_{\vec{T}_{2k} \in C_2} \frac{\vec{T}_{1j} \times (\vec{T}_{2k} \times \vec{r}_{kj})}{r_{kj}^3} \quad (6)$$

where $F_d$ is the force on wire1 from wire2. Here both wire1 and wire2 are in discrete form, which consist of discrete current element vectors respectively. $C_1$ and $C_2$ are the two sets of discrete current elements for wire1 and wire2 respectively. In another word, all the discrete vectors in $C_1$ constitute the discrete form of wire1, and all the discrete vectors in $C_2$ constitute the discrete form of wire2. $T_{1j}$ and $T_{2k}$ are the current element vectors in $C_1$ and $C_2$ respectively. $r_{kj}$ is the vector from $T_{2k}$ to $T_{1j}$.

Some examples of discretized current-carrying wires are shown in Fig. 3. For display in the figures, the continuous current directions are discretized into 8 directions: {east, west, north, south, northeast, northwest, southeast, southwest}. Fig. 3(a) shows the discrete form of a continuous current with the rectangle shape. Fig. 3(b) shows the discrete form of a continuous current with the circle shape. Fig. 3(c) shows the discrete form of a continuous current with the straight-line shape.

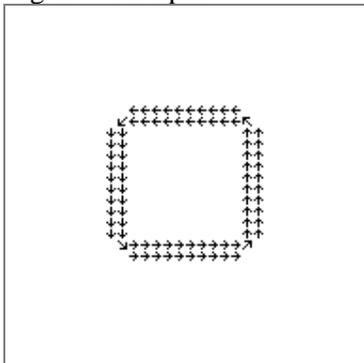

**Fig. 3(a) The discrete form of a continuous rectangle current**

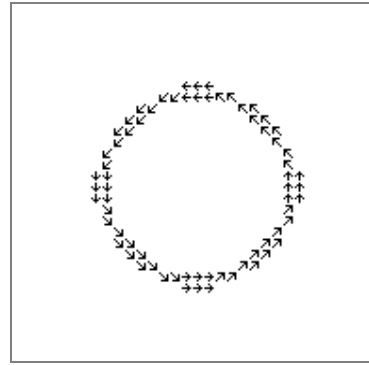

**Fig. 3(b) The discrete form of a continuous circle current**

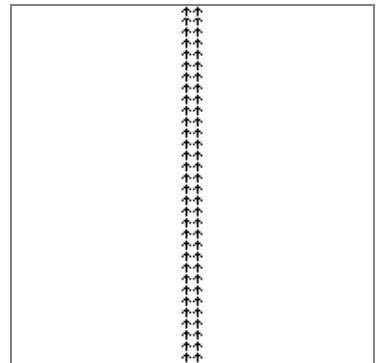

**Fig. 3(c) The discrete form of a continuous straight-line current**

In further simulation experiments, the force is calculated between the original discrete current and the one with a certain displacement by a shifting transformation. The results are shown in Fig. 4, where $C_1$ and $C_2$ are two discrete currents and $F_d$ is the force on $C_1$ applied by $C_2$. The virtual force on $C_1$ by $C_2$ is calculated according to Equation (6). In the simulation experiments, the discrete current $C_2$ attracts its shifted version $C_1$, which has the effect of restoring the shifted one back to the original position. This phenomenon inspires the matching approach by virtual electromagnetic interaction between images, which is discussed in the following sections.

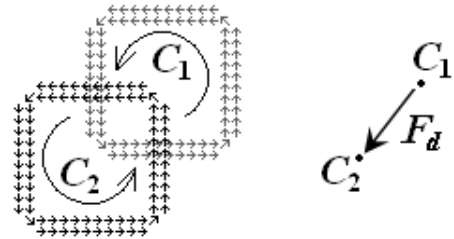

**Fig. 4(a) The simulation of the interaction between two discrete rectangle current**

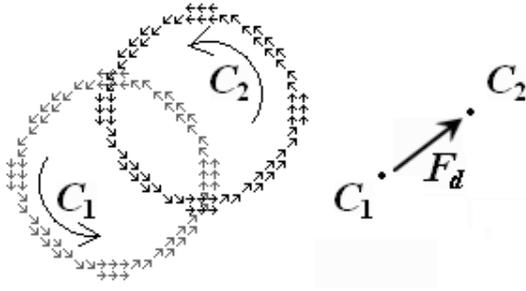

**Fig. 4(b) The simulation of the interaction between two discrete circle current**

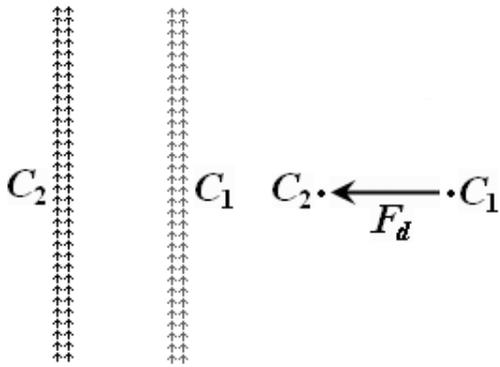

**Fig. 4(c) The simulation of the interaction between two discrete straight-line current**

## 3 Electromagnetism-inspired Image Matching for Shifting Transformation

Among the current methods of image registration, feature-based methods have the notable advantage of much less computation load. There are several important image features which can represent image structure, and edge is one of them. On the other hand, many physical experiments and theoretical analysis indicate that the electro-magnetic interaction can cause a shifted current-carrying wire attracted to another wire of the same shape on the original location (Strictly speaking, the attraction between the original wire and the shifted one exists only in a certain range of displacement). The simulation experiments in the above section also produce such results. If a proper line feature representing the image structure can be found in the image, it is possible to exploit the above location-restoring effect of electro-magnetic interaction for image matching. The edge feature is just suitable for this requirement.

### 3.1 The significant edge current

In this paper, the "significant edge points" are extracted for image matching. The significant edges are definite borders of regions, and there is sharp change of grayscale across the significant edge lines. In the proposed method, first the significant edge points are extracted in the two images to be matched. The virtual currents in the image are defined as the set of discrete current elements on the significant edge lines. Then the interaction force between the virtual currents in the two images is calculated, which inspires a novel matching approach.

Canny operator is widely used to extract edge lines [20,21]. To improve computation efficiency, in this paper a simplified Canny-like method is proposed to extract the significant edges in image. First, Sobel operator is used to estimate the gradient fields of the two images to be matched. Then the thresholding process of the gradient magnitude is performed to extract the definite edge points, in which only the points with a magnitude larger than the threshold value are reserved for further process. (The threshold value is set as a predefined percent of the maximum value of gradient magnitude.) After that, "non-maximum suppression" is performed to get thin edge lines. For an edge point, it is reserved as a significant edge point only if its gradient magnitude is larger than the adjacent points on at least two of the following pairs of directions: west and east, north and south, northwest and southeast, northeast and southwest.

With the above process, the significant edge lines can be extracted. However, the direction of current element should be along the tangent direction of the wire, while the gradient vector is perpendicular to the tangent direction of the edge line. Therefore, all the gradient vectors on the significant edge lines rotate 90 degrees to form the virtual edge current. Then the vector direction after rotation is along the tangent direction of edge lines. In another word, on the significant edge lines, the virtual current elements are obtained by rotating the gradient vectors 90 degrees counterclockwise. For a clear demonstration, a simple example is shown in Fig. 5. All the virtual current element vectors on the significant edges form the virtual currents in the image, i.e. the virtual edge current in an image is a set of discrete current element vectors on the significant edge lines.

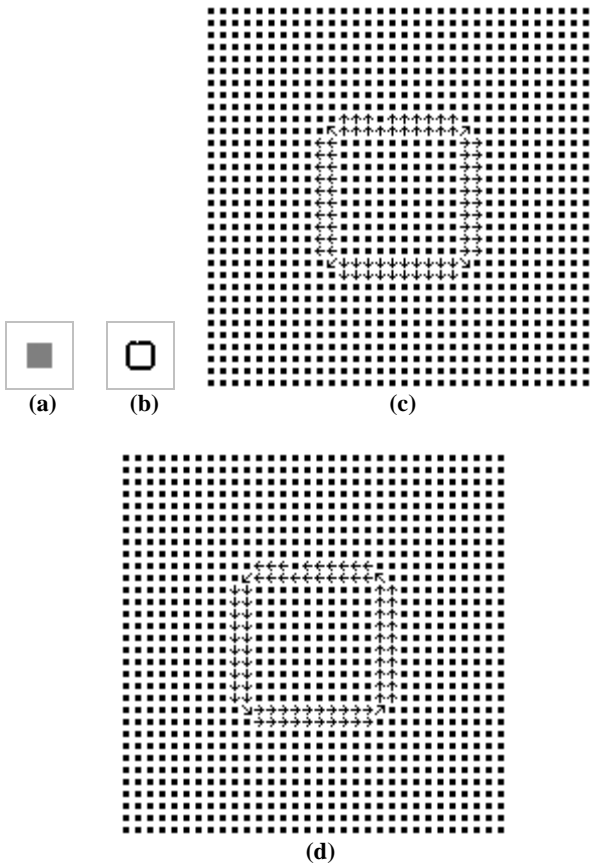

**Fig. 5 An example of the significant edge lines and the corresponding virtual current of a square shape image**
(a) The image of the square shape
(b) The significant edge lines of (a)
(c) The gradient vectors on the significant edge lines
(d) The discrete virtual current

Fig. 5(a) shows a simple image of a square. Fig. 5(b) shows its significant edge lines extracted. Fig. 5(c) shows the direction distribution of the discrete gradient vectors, where the continuous gradient direction is discretized into eight directions for display: {east, west, north, south, northeast, northwest, southeast, southwest}. Fig. 5(d) shows the direction distribution of the discrete current elements, which is the rotated version of gradient vector. The dots in Fig. 5(c) and Fig. 5(d) show the points with no current elements.

Some other examples of extracting virtual edge current are shown from Fig. 6 to Fig. 10. Fig. 6(a) to Fig. 9(a) show some simple images of the size $32 \times 32$. Some real world images of the size $128 \times 128$ are shown in Fig. 10. The significant edge lines extracted are also shown, which will be used as virtual edge currents in the following matching process.

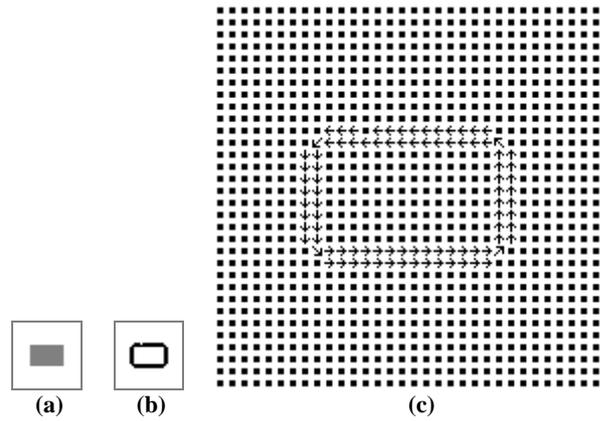

**Fig. 6 The significant edge lines and the corresponding virtual current of a rectangle shape image**
(a) The image of the rectangle shape
(b) The significant edge lines of (a)
(c) The discrete virtual current

Fig. 6(a) shows an image of a rectangle. Fig. 6(b) shows its significant edge lines. Fig. 6 (c) shows the virtual current elements on the significant edge lines, where the arrows show the discrete directions of discrete current elements. The dots in Fig. 6(c) show the points with no current elements.

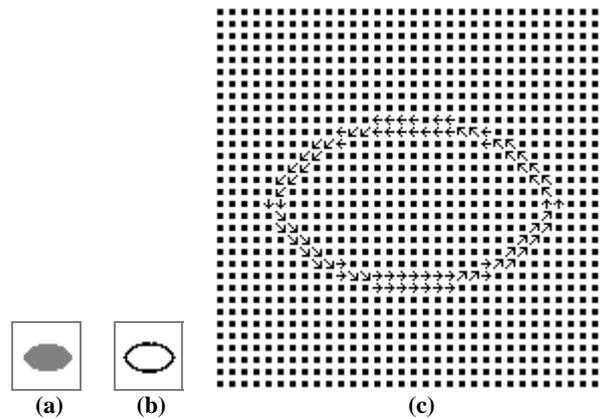

**Fig. 7 The significant edge lines and the corresponding virtual current of an ellipse shape image**
(a) The image of the ellipse shape
(b) The significant edge lines of (a)
(c) The discrete virtual current

Fig. 7(a) shows an image of an ellipse. Fig. 7(b) shows its significant edge lines. Fig. 7(c) shows the virtual current elements on the significant edge lines, where the arrows show the discrete directions of discrete current elements.

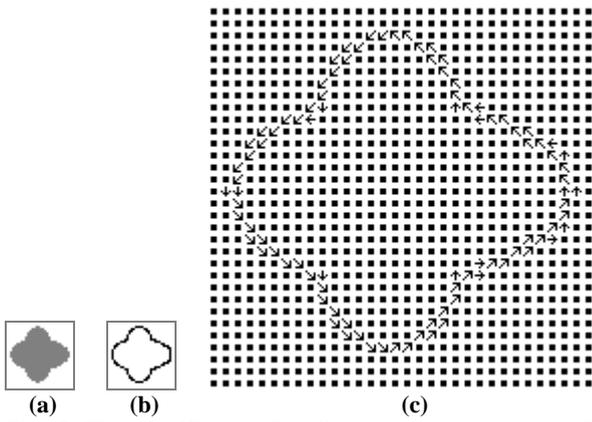

**Fig. 8 The significant edge lines and the corresponding virtual current of an irregular shape image**
**(a) The image of the irregular shape**
**(b) The significant edge lines of (a)**
**(c) The discrete virtual current**

Fig. 8(a) shows an image of an irregular shape. Fig. 8(b) shows its significant edge lines. Fig. 8(c) shows the virtual current elements on the significant edge lines, where the arrows show the discrete directions of discrete current elements.

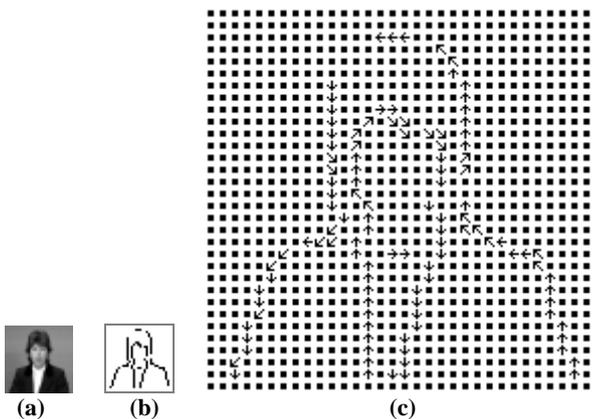

**Fig. 9 The significant edge lines and the corresponding virtual current of a shrunken image of broadcaster**
**(a) The shrunken broadcaster image**
**(b) The significant edge lines of (a)**
**(c) The discrete virtual current**

Fig. 9(a) shows the shrunken version of a broadcaster image. Fig. 9(b) shows its significant edge lines. Fig. 9(c) shows the virtual current elements on the significant edge lines, where the arrows show the discrete directions of discrete current elements.

In Fig. 10, the real world images and their significant edge lines are shown, including the images of the broadcaster, the cameraman, the peppers, the locomotive, the boat, the house and a medical image of brain.

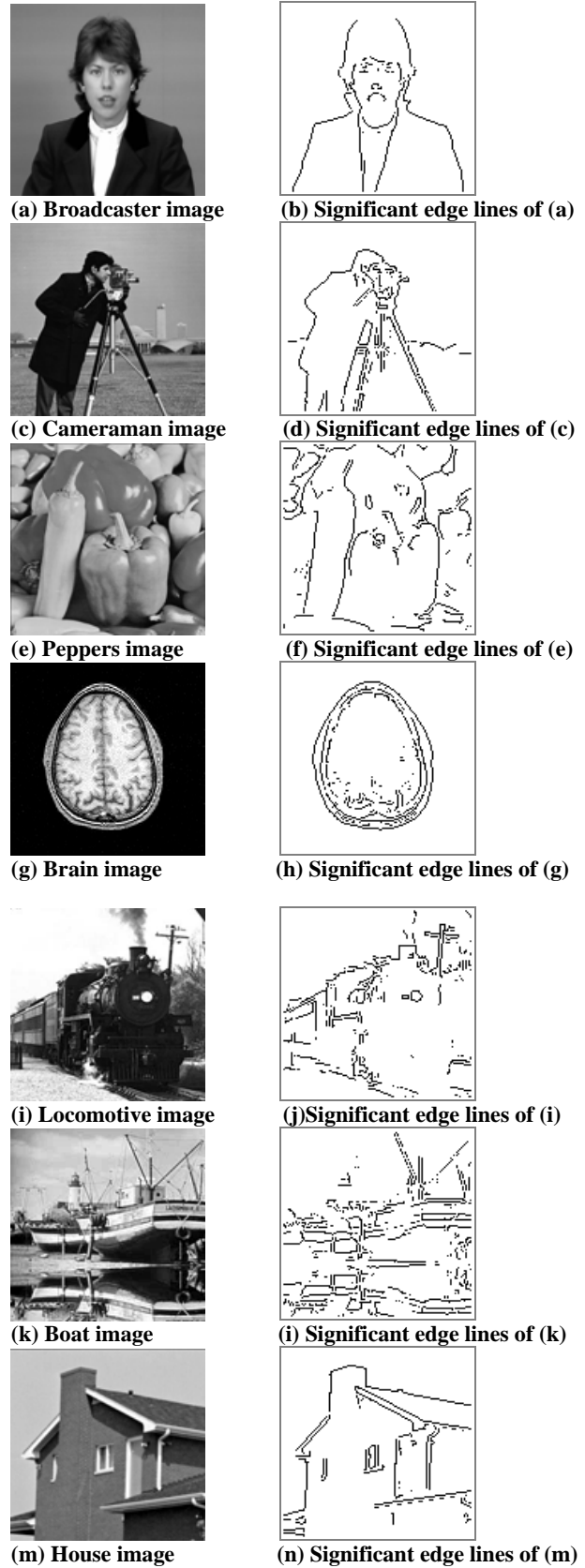

**(a) Broadcaster image**    **(b) Significant edge lines of (a)**
**(c) Cameraman image**    **(d) Significant edge lines of (c)**
**(e) Peppers image**    **(f) Significant edge lines of (e)**
**(g) Brain image**    **(h) Significant edge lines of (g)**
**(i) Locomotive image**    **(j) Significant edge lines of (i)**
**(k) Boat image**    **(i) Significant edge lines of (k)**
**(m) House image**    **(n) Significant edge lines of (m)**

**Fig. 10 A group of real world images and their significant edge lines**

## 3.2 The interaction of the virtual edge currents between two related images

If the virtual currents in two images are extracted respectively, the interaction force between them can be calculated according to Equation (6). Suppose the sets of discrete current elements in image1 and image2 are $C_1$ and $C_2$ respectively. Each current element $T_{1j}$ in $C_1$ is applied the force by all the current elements in $C_2$:

$$\vec{F}_{1j} = A \cdot \sum_{\vec{T}_{2k} \in C_2} \frac{\vec{T}_{1j} \times (\vec{T}_{2k} \times \vec{r}_{kj})}{r_{kj}^3} \qquad (7)$$

where $F_{1j}$ is the force on $T_{1j}$ from $C_2$. $T_{2k}$ is a current element vector in $C_2$. $r_{kj}$ is the vector from $T_{2k}$ to $T_{1j}$. $A$ is a predefined constant value for programming convenience. Some simulation examples of the force on the virtual current elements are shown in Fig. 11 to Fig. 13. (Here suppose the two images are on the same plane.)

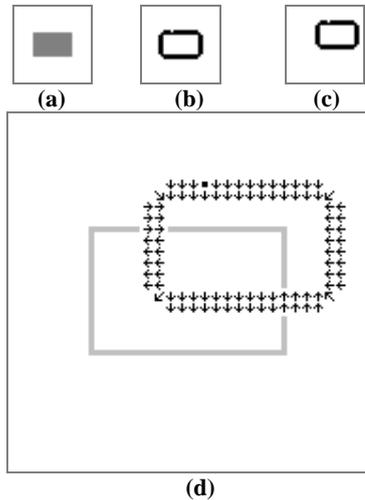

**Fig. 11 The virtual force on each current element in the shifted image for the rectangle image**
(a) the rectangle image
(b) the significant edge lines of (a)
(c) the significant edge lines of the shifted image
(d) the forces on the discrete current elements in the shifted image

Fig. 11(a) shows the image of a rectangle. The significant edge lines of the original image and the shifted one are shown in Fig. 11(b) and Fig. 11(c). The translation from (b) to (c) on *x* and *y* coordinates are 5 and -4 respectively (in the screen coordinates). Fig. 11(d) shows the force direction on each current element in the shifted image, which is applied by the virtual current in the original image. The gray lines in Fig. 11(d) show the original position of the rectangle. The force on each current element in the shifted image is calculated according to Equation (7).

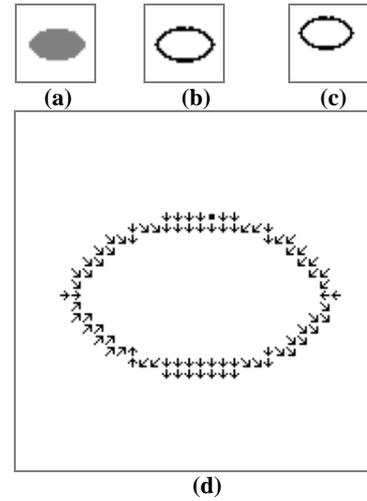

**Fig. 12 The virtual force on each current element in the shifted image for the ellipse image**
(a) the ellipse image
(b) the significant edge lines of (a)
(c) the significant edge lines of the shifted image
(d) the forces on the discrete current elements in the shifted image

Fig. 12(a) shows the image of an ellipse. The significant edge lines of the original image and the shifted one are shown in Fig. 12(b) and Fig. 12(c). The translation from (b) to (c) on *x* and *y* coordinates are -6 and -6 respectively. Fig. 12(d) shows the force direction on each current element in the shifted image, which is applied by the virtual current in the original image.

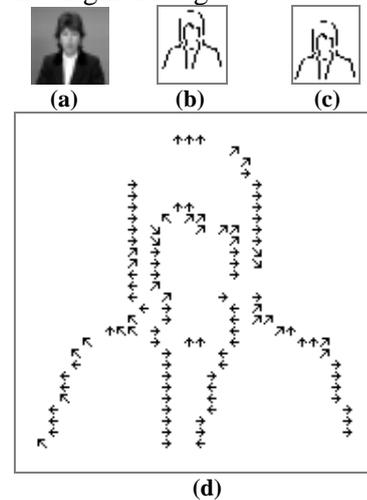

**Fig. 13 The virtual force on each current element in the shifted image for the shrunken image of broadcaster**
(a) the shrunken image of the broadcaster
(b) the significant edge lines of (a)
(c) the significant edge lines of the shifted image
(d) the forces on the discrete current elements in the shifted image

Fig. 13(a) shows the shrunken image of the broadcaster. The significant edge lines of the original image and the shifted one are shown in Fig. 13(b) and Fig. 13(c). The translation from (b) to (c) on *x* and *y* coordinates are -1 and 5 respectively. Fig. 13(d) shows the force direction on each current element in the shifted image, which is applied by the virtual current in the original image.

It is interesting and worthwhile to investigate the total force on the shifted image by the original one. For the whole set $C_1$ as the virtual current in image1, it is applied the force from $C_2$. The total force on $C_1$ from $C_2$ is the summation of the force on each of its component $T_{1j}$:

$$\vec{F}_{12} = A \cdot \sum_{\vec{T}_{1j} \in C_1} \sum_{\vec{T}_{2k} \in C_2} \frac{\vec{T}_{1j} \times (\vec{T}_{2k} \times \vec{r}_{kj})}{r_{kj}^3} \quad (8)$$

where $F_{12}$ is the force on $C_1$ from $C_2$. $T_{1j}$ and $T_{2k}$ are the current element vectors in $C_1$ and $C_2$ respectively. $r_{kj}$ is the vector from $T_{2k}$ to $T_{1j}$.

Assume two images have large matched part but there is shifting transformation between them. Their virtual edge currents are almost consistent due to the large matched parts. Therefore, the interaction between their virtual edge currents $C_1$ and $C_2$ will be similar to that between the two current-carrying wires with the same shape, with which one wire will attract the other to coincide in position. Such effect of attraction will make the shifted wire restore its original position, which can serve as a novel way for image matching with shifting transformation.

Here the direction of the total force is investigated experimentally. The total forces in Fig. 11(d), Fig. 12(d) and Fig. 13(d) are calculated according to Equation (8). The results are shown as following.

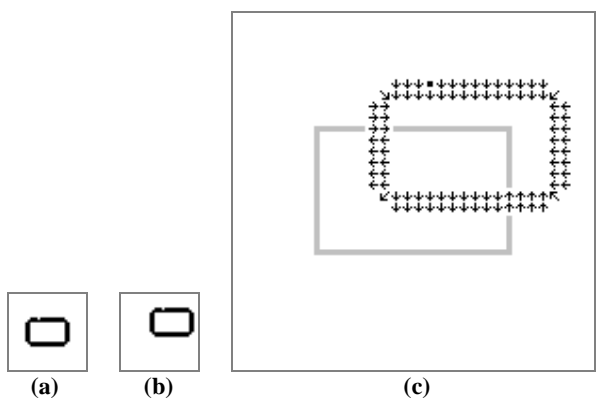

**Fig. 14 The significant edge lines of the original and shifted rectangle images, and the forces on the shifted image**
(a) the significant edge lines of the rectangle image
(b) the significant edge lines of the shifted image
(c) the forces on the discrete current elements in the shifted image

Fig. 14(b) moves northeast relative to Fig. 14(a). It can be found in Fig. 14(c) that most forces along the horizontal direction are westward, and most forces along the vertical direction are southward. After calculating the total force by programming, the total force on Fig. 14(b) applied by Fig. 14(a) has such direction: its *x* component is negative and its *y* component is positive. Under the screen coordinate shown in Fig. 15, the effect of the total force is making the shifted image move southwest. In another word, in this experiment the effect of the total force is restoring the shifted image back to its original position.

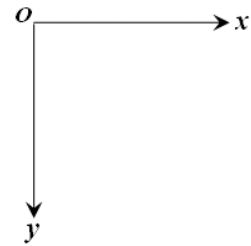

**Fig. 15 The screen coordinates**

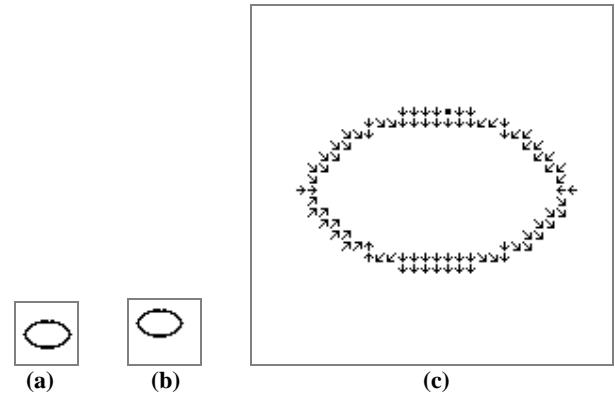

**Fig. 16 The significant edge lines of the original and shifted ellipse images, and the forces on the shifted**
(a) the significant edge lines of the ellipse image
(b) the significant edge lines of the shifted image
(c) the forces on the discrete current elements in the shifted image

In another experiment, Fig. 16(b) moves northwest relative to Fig. 16(a). The total force on Fig. 16(b) has such direction: its *x* component and *y* component are both positive. Under the screen coordinate, the effect of the total force is making the shifted image move southeast. In another word, in this experiment, the effect of the total force is also restoring the shifted image back to its original position.

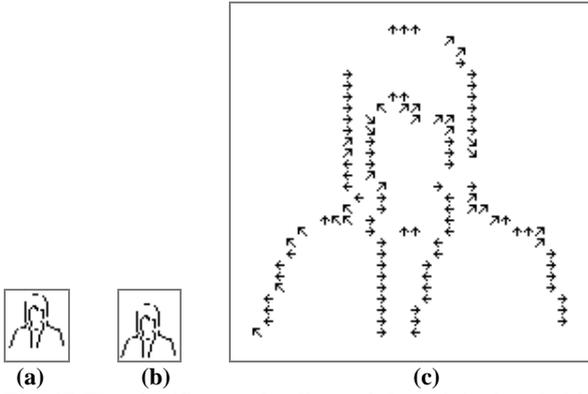

(a)  (b)  (c)

**Fig. 17 The significant edge lines of the original and shifted broadcaster images (the shrunken version), and the forces on the shifted image**
(a) the significant edge lines of the shrunken image of the broadcaster
(b) the significant edge lines of the shifted image
(c) the forces on the discrete current elements in the shifted image

In the experiment of the broadcaster image, Fig. 17(b) moves southwest relative to Fig. 17(a). The total force on Fig. 17(b) has such direction obtained by calculation: its *x* component is positive and its *y* component is negative. Under the screen coordinate, the effect of the total force is making the shifted image move northeast. In another word, in this example the effect of the total force is still restoring the shifted image back to its original position. Other experiments for simple images also demonstrate the same phenomenon as shown in the above results, which inspires the matching approach in the paper.

## 3.3 Matching for shift transformation by virtual electromagnetic interaction

In the above simulation experiments, one image is the shifted version of the other, and the total force between them has the effect of restoring the shifted image back to the original position. This inspires a novel matching method for shift transformation.

### 3.3.1 The distribution map of total force

Suppose there are two images: image2 is the original one and image1 is the shifted version of image2. It is interesting and meaningful to investigate whether the "position restoring" effect exists for arbitrary shift of image position. In order to investigate the total force on image1 with different shift of position, the experiment is designed as following: image2 is fixed in position, and the center of image1 is moved onto each discrete position on the image plane (the two images are on the same plane). Calculate the total force on image1 by image2 for each shift. Then the distribution map of force direction for different shift can be drawn. The total force distribution map shows the total force on each position which the center of image1 can be shifted to. Fig. 18 to Fig. 20 show some experiment results for the test images of the size $32 \times 32$.

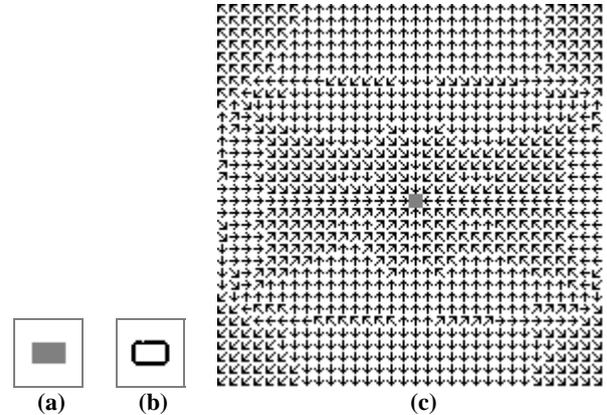

(a)  (b)  (c)

**Fig. 18 The direction distribution map of the total force for the rectangle image (the gray point is the original position before shifting)**
(a) the rectangle image
(b) the significant edge lines of (a)
(c) the direction distribution map of the total force

Fig. 18(a) shows the image of a rectangle. Fig. 18(b) shows its significant edge lines. Fig. 18(c) shows the direction distribution map of the total force. There is an arrow on each discrete position in Fig. 18(c).The direction of the arrow shows the direction of the total force on image1 applied by image2 when the center of image1 moves onto that position. The continuous force directions are discretized into 8 directions for display: {east, west, north, south, northeast, northwest, southeast, southwest}. The gray dot at the center represents the original position of the image center before shifting (the center of image2).

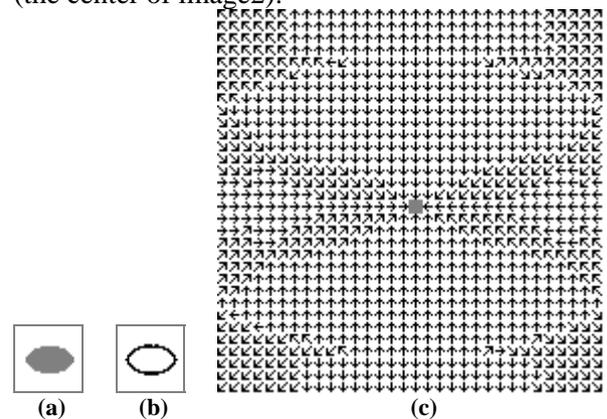

(a)  (b)  (c)

**Fig. 19 The direction distribution map of the total force for the ellipse image (the gray point is the original position before shifting)**
(a) the ellipse image
(b) the significant edge lines of (a)
(c) the direction distribution map of the total force

Fig. 19(a) shows the image of an ellipse. Fig. 19(b) shows its significant edge lines. Fig. 19(c) shows the direction distribution of the total force. The gray dot at the center represents the original position of image center before shifting (the center of image2). The arrow on each discrete position represents the discrete direction of the total force on image1 applied by image2 when the center of image1 moves onto that position.

In Fig 18(c) and Fig. 19(c), most of the force vectors around the original position (the gray point in the figures) point to the center (i.e. the original position), which form a force field area attracting the shifted image back to the original position. However, in the border areas of Fig 18(c) and Fig. 19(c), the force vectors point to the opposite trend, which do not have the "position restoring" effect for shift transformation. Therefore, it is necessary to study the effective range of shifting within which the virtual force can bring the shifted image back to the original position.

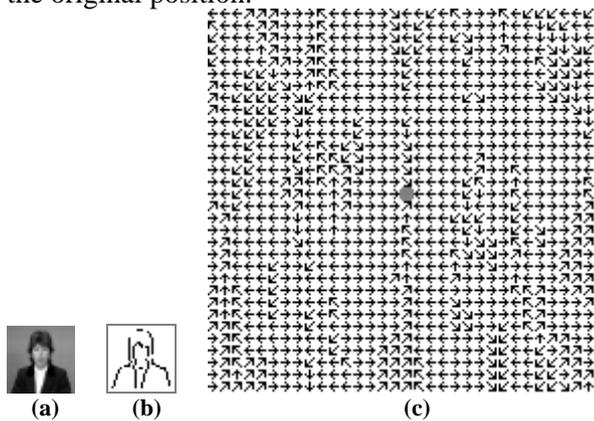

**Fig. 20 The direction distribution map of the total force for the shrunken image of the broadcaster (the gray point is the original position before shifting)**
(a) the shrunken image of the broadcaster
(b) the significant edge lines of (a)
(c) the direction distribution map of the total force

Fig. 20(a) shows the shrunken image of the broadcaster. Fig. 20(b) shows its significant edge lines. Fig. 20(c) shows the direction distribution of the total force. The gray dot at the center represents the original position of image center before shifting (the center of image2). The arrow on each discrete position represents the discrete direction of the total force on image1 applied by image2 when the center of image1 moves onto that position.

### 3.3.2 The convergence and divergence points in the distribution map of total force

Since the distribution map of the total force can be obtained such as Fig. 18(c), Fig. 19(c) and Fig. 20(c), it is possible to move the shifted image1 towards the original position according to the force distribution map, which is an interesting and meaningful topic. If image1 (with a certain displacement in position) can be guided back to the original position step by step, a novel matching approach for shift transformation can be implemented.

Therefore, experiments are designed and carried out by programming for the above topic. For each discrete position ($x,y$) (i.e. shifted image center), there is a total force. According to the direction of the total force, move to the next position. Then repeat the moving step by step in the force distribution map. If the original position of image center (the center of image2) is finally arrived, a path will be formed along which the shifted image1 can be guided back to the original position. In another word, the matching for image1 and image2 can be implemented. In the experiment, each position is taken as the starting point of the above moving process, and the final result of moving in the force distribution map is recorded and classified.

In Fig. 18(c) and Fig. 19(c), two different cases of the moving resultscan be found. In the first case, the path finally ends at the original center position, which leads to a successful matching. The starting positions in such case are defined as the *convergence points* in the force distribution map. In the experiments, for the convergence points, when the path arrives at the original center point, the moving process can be ended naturally because the next moving steps will oscillate back and forth between the original center point and its adjacent point. Such points like the original center point are defined as the *balance points* in the force distribution map, from which the next moving steps will oscillate. If such oscillation is found in the experiments, the moving process can be ended. An example of convergence point found in Fig. 18(c) is shown in Fig. 21.

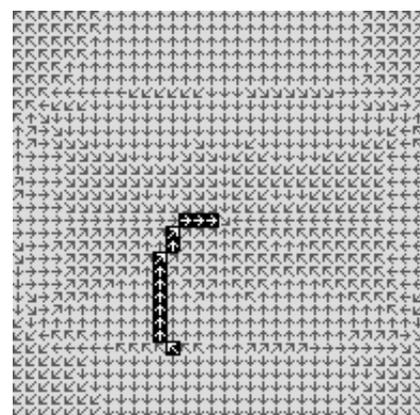

**Fig. 21 An example of convergence point found in Fig. 18(c)**

Another example of convergence point in Fig. 19(c) is shown in Fig. 22.

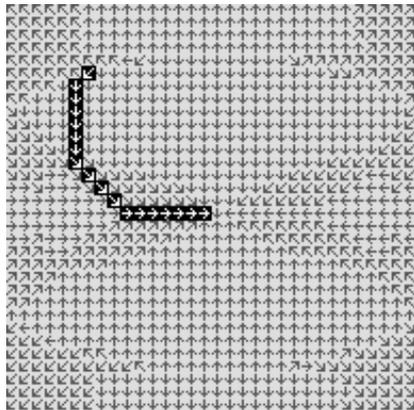

**Fig. 22 Another example of convergence point found in Fig. 19(c)**

Therefore, if the center of image1 is shifted onto a position of convergence point, image1 can be moved back to the original position (i.e. the position of image2) according to the guidance of the force distribution map.

The second case found in the experiments is quite different from the first one. For some starting points in the force distribution map, when moving according to the force direction step by step, the path will finally exceed the range of the image area and never return back to the original center position. In such case, the matching for shift transformation can not be accomplished. The starting points in such case are defined as the *divergence points*. An example of the divergence points found in Fig. 18(c) is shown in Fig. 23.

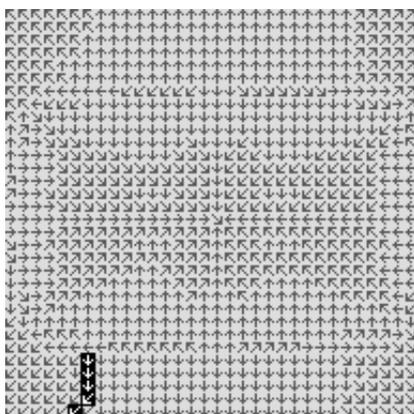

**Fig. 23 An example of the divergence points found in Fig. 18(c)**

Another example of the divergence point in Fig. 19(c) is shown in Fig. 24.

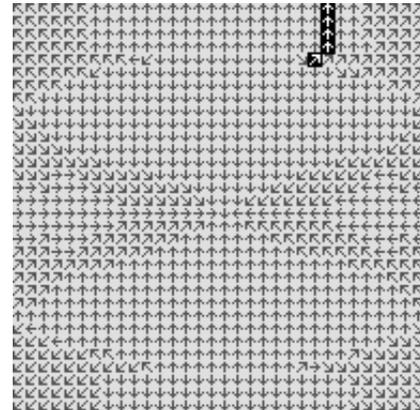

**Fig. 24 Another example of the divergence points found in Fig. 19(c)**

### 3.3.3 The classification of the positions in the distribution map of total force

For Fig. 18(c) and Fig. 19(c), investigate each position in the force distribution map as starting point, and each position can be determined as a convergence or divergence point. Therefore, a classification of the positions in the direction distribution can be obtained. Fig. 25 and Fig. 26 show the classification result of Fig. 18(c) and Fig. 19(c). In Fig. 25(d) and Fig. 26(d), the black area represents the convergence points, and the white area represents the divergence points.

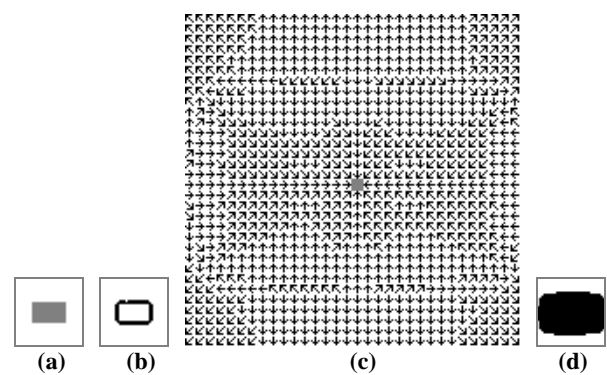

**(a)**    **(b)**    **(c)**    **(d)**

**Fig. 25 The classification result of the total force distribution for the rectangle image**
**(a) the rectangle image**
**(b) the significant edge lines of (a)**
**(c) the direction distribution map of the total force**
**(d) the classification of the direction distribution (the black area is the convergence area for matching)**

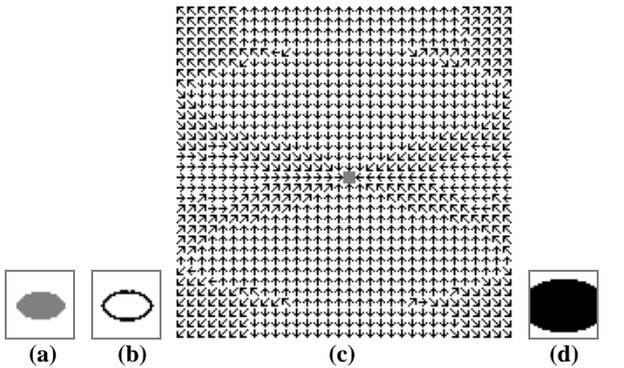

**Fig. 26 The classification result of the total force distribution for the ellipse image**
(a) the ellipse image
(b) the significant edge lines of (a)
(c) the direction distribution map of the total force
(d) the classification of the direction distribution (the black area is the convergence area for matching)

In Fig. 25(d) and Fig. 26(d), if the center of image1 is shifted to a position within the black area of convergence points, image1 can be moved back to the original position (i.e. the position of image2) according to the guidance of the force distribution. Therefore, the segmentation of the force distribution is useful in the analysis of the effectiveness of matching by virtual electromagnetic interaction between images.

For real world images which are more complex than the simple test images, another case is found in the experiments besides the convergence and divergence points. For some starting points, the moving process guided by the total force is trapped into a local balance point rather than arrive at the original image center. In another word, the moving finally becomes oscillating back and forth around the local balance point. Such points are defined as the *locally trapped points*. In Fig. 20(c) such case can be found. An example is shown in Fig. 27, in which the moving process reaches a local balance point. For the locally trapped points, the matching can not be accomplished.

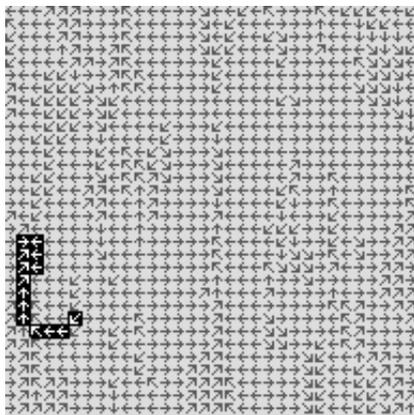

**Fig. 27 An example of the locally trapped points found in Fig. 20(c)**

Therefore, it is indicated by the experiments that there are 3 kinds of points in the direction distribution map: convergence points, divergence points and locally trapped points. After each discrete point is classified as one of the above 3 categories, a classification result can be obtained, which is the "classification map". For an image in matching, the force distribution can be calculated and the classification map can be obtained. The classification map of Fig. 20(c) is shown in Fig. 28. In Fig. 28(d), the black points are the convergence points, the white points are the divergence points, and the grey points represent the locally trapped points. The effectiveness of the matching method by virtual electromagnetic interaction can be clearly and directly perceived in the classification map. Only if the center of the shifted image is in the area of convergence points, the virtual force can guide it back to the original position and accomplish the matching.

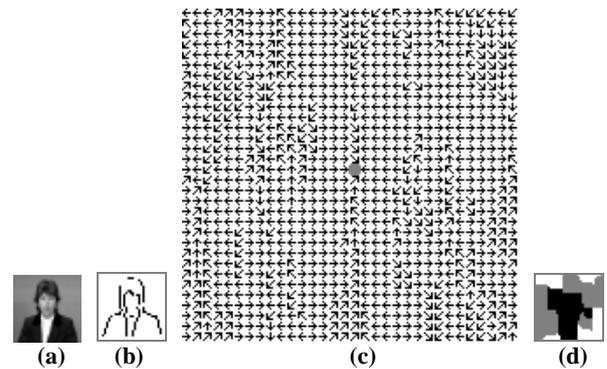

**Fig. 28 The classification map of the total force distribution for the shrunken image of the broadcaster**
(a) the shrunken image of the broadcaster
(b) the significant edge lines of (a)
(c) the direction distribution map of the total force
(d) the classification map of the direction distribution (the black area contains the convergence points for matching, and the gray areas represent locally trapped points)

Therefore, if the area of convergence point gets larger, the matching by virtual electromagnetic interaction will be more effective. In another experiment, the classification map with the broadcaster image of the size $128 \times 128$ is obtained, which is shown in Fig. 29. In Fig. 29(b), the black, white and gray points represent the convergence points, the divergence points, and the locally trapped points respectively. In Fig. 29(b), the convergence point area is not large enough for matching with large shift between images. Some possible improvement is studied in the following section.

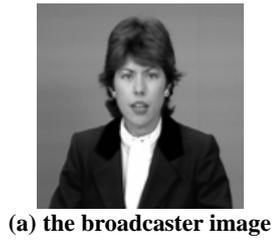

**(a) the broadcaster image**

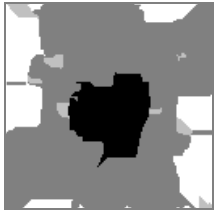

**(b) the classification map of the total force distribution**

**Fig. 29 The classification map for the broadcaster image**

### 3.3.4 Improve the matching effectiveness in 3D space

In the above experiments, the shifted image (image1) and the original image (image2) are on the same plane. From the view point of 3D space, image1 only has displacements on $x$ and $y$ coordinates. Its height on $z$-coordinate is the same as image2. In order to improve the effectiveness of the proposed approach, the authors have attempted to place the two images on different heights (i.e. different $z$ coordinates in the 3D space), and studied the changes in the classification map. In another word, image1 and image2 are placed on two different but parallel planes in 3D space, and image1 is the shifted one on $x$ and $y$ coordinates with respect to image2.

In the following experiments, the two images are placed on two parallel planes at different heights respectively in 3D space. Here "pixel" is used as the unit to measure the height in 3D space. Then the virtual force vectors on image1 applied by image2 will have $x$, $y$ and $z$ components. Here only the $x$ and $y$ components of the force are considered to calculate the total force on the $x$-$y$ plane. Some of the experiment results for the test images with the size $32 \times 32$ are shown in Fig. 30 to Fig. 33. In the results, the black points represent the convergence points for matching.

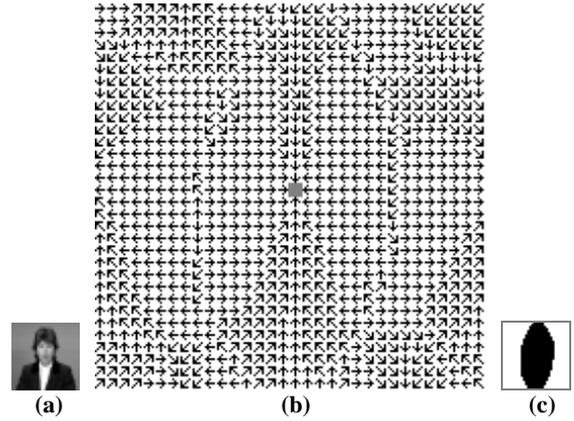

**Fig. 30 The classification map of total force for the shrunken image of the broadcaster (the distance between the images is 5 pixels)**
**(a) the shrunken image of the broadcaster**
**(b) the direction distribution map of total force (the distance between the images is 5)**
**(c) the classification map (the distance between the images is 5)**

Fig. 30(a) shows the shrunken image of the broadcaster. The distance between the original image and the shifted one is 5 pixels (on $z$ coordinate). Fig. 30(b) shows the direction distribution of total force. Fig. 30(c) shows the classification map for Fig. 30(b). Compared to Fig. 28(d), the convergence point area in Fig. 30(c) is larger, and the area of locally trapped points disappears in Fig. 30(c).

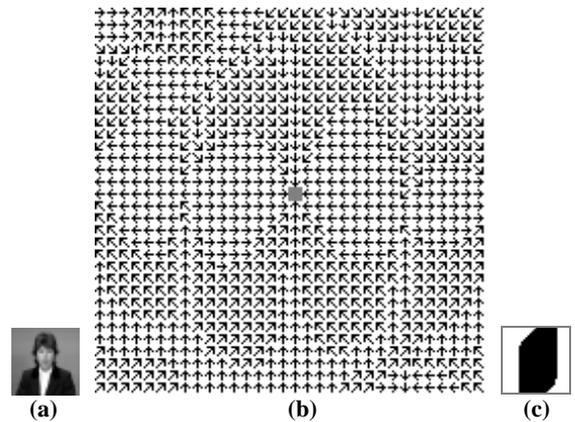

**Fig. 31 The classification map of total force for the shrunken image of the broadcaster (the distance between the images is 8 pixels)**
**(a) the shrunken image of the broadcaster**
**(b) the direction distribution map of total force (the distance between the images is 8)**
**(c) the classification map (the distance between the images is 8)**

The total force distribution and the corresponding classification map are changed with the increasing of the distance between the two images. Fig. 31 shows another result by further increasing the distance between the original image and the shifted one. The distance between the original image and the shifted one is 8 pixels (on *z* coordinate). Fig. 31(b) shows the direction distribution of total force. Fig. 31(c) shows the classification map for Fig. 31(b). Compared to Fig. 30(c), the convergence point area in Fig. 31(c) becomes larger, which indicates the effectiveness of matching by virtual electromagnetic interaction is further improved.

Another two experiment results are shown in Fig. 32 and Fig. 33. These two are for the test images in Fig. 25 and Fig. 26. The distance between the original image and the shifted one is 8 pixels (on *z* coordinate). The original images, the force distribution and the classification map are shown respectively. In Fig. 32(c) and Fig. 33(c), the black and white points represent the convergence and divergence points respectively. Compared to Fig. 25(d) and Fig. 26(d), Fig. 32(c) and Fig. 33(c) indicate obvious improvement in the increasing of convergence point area. It is proved by experiments that setting proper distance between the two images (on *z* coordinate) will improve the effectiveness of the matching approach by virtual electromagnetic interaction.

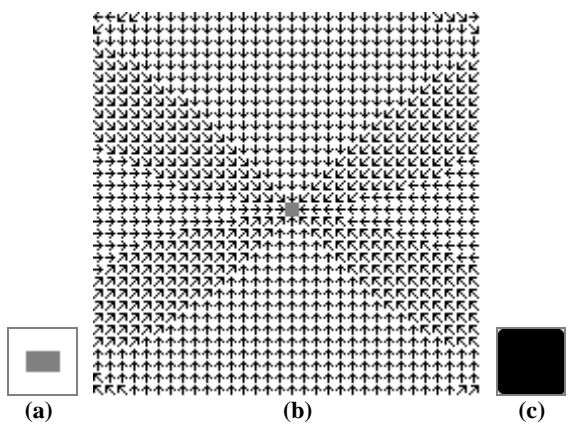

(a)      (b)      (c)

**Fig. 32 The classification map of total force for the rectangle image (the distance between the images is 8 pixels)**
**(a) the rectangle image**
**(b) the direction distribution map of total force (the distance between the images is 8)**
**(c) the classification map (the distance between the images is 8)**

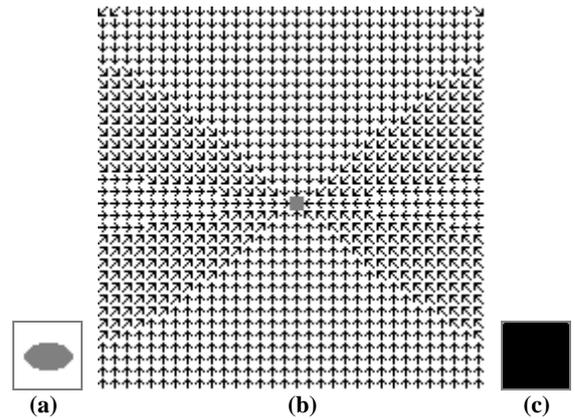

(a)      (b)      (c)

**Fig. 33 The classification map of total force for the ellipse image (the distance between the images is 8 pixels)**
**(a) the ellipse image**
**(b) the direction distribution map of total force (the distance between the images is 8)**
**(c) the classification map (the distance between the images is 8)**

Experiments are also carried out for real world images of the size $128 \times 128$. Some of the results are shown in Fig. 34 to Fig. 39. The original images are shown together with the classification maps under different height distances between the two images. In the classification maps, the black, white and gray points represent the convergence points, the divergence points, and the locally trapped points respectively. The experimental results again prove the advantage of setting proper height distance between the two images to be matched. In the experiments, with the increasing of the height distance, the convergence point area also increases. Especially, in Fig. 35(f) and Fig. 36(f) the convergence point areas almost cover the whole image region.

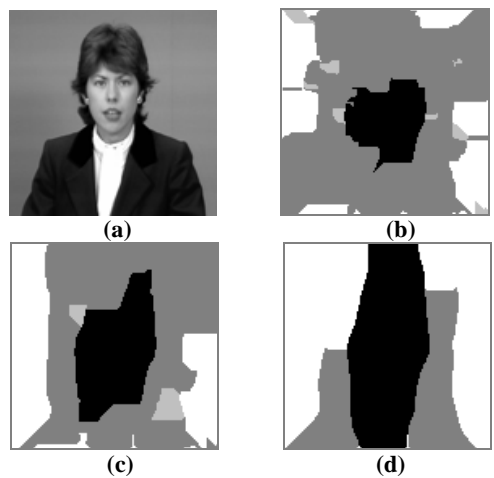

(a)      (b)

(c)      (d)

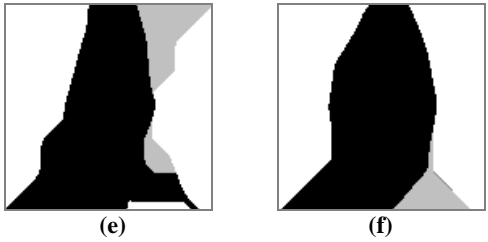

**Fig. 34** The classification maps for the broadcaster image with various height distances between the two images in 3D space
(a) the broadcaster image
(b) the classification map of total force direction (the distance between the images is 0)
(c) the classification map of total force direction (the distance between the images is 1)
(d) the classification map of total force direction (the distance between the images is 4)
(e) the classification map of total force direction (the distance between the images is 8)
(f) the classification map of total force direction (the distance between the images is 16)

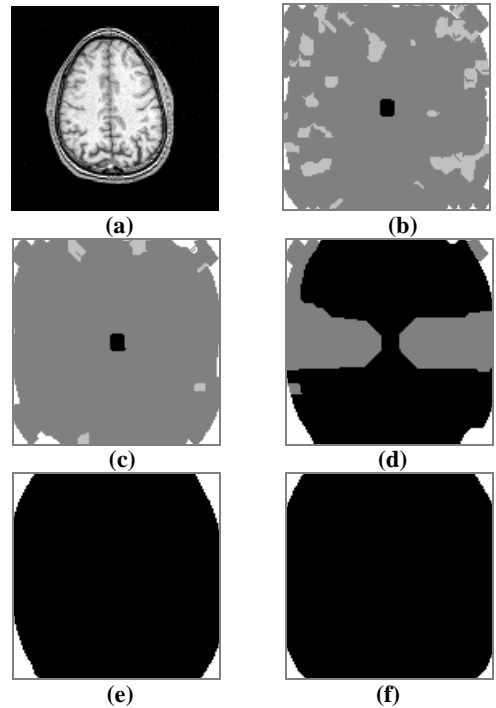

**Fig. 36** The classification maps for the brain image with various height distances between two images in 3D space
(a) the medical image of the brain
(b) the classification map of total force direction (the distance between the images is 0)
(c) the classification map of total force direction (the distance between the images is 1)
(d) the classification map of total force direction (the distance between the images is 2)
(e) the classification map of total force direction (the distance between the images is 4)
(f) the classification map of total force direction (the distance between the images is 8)

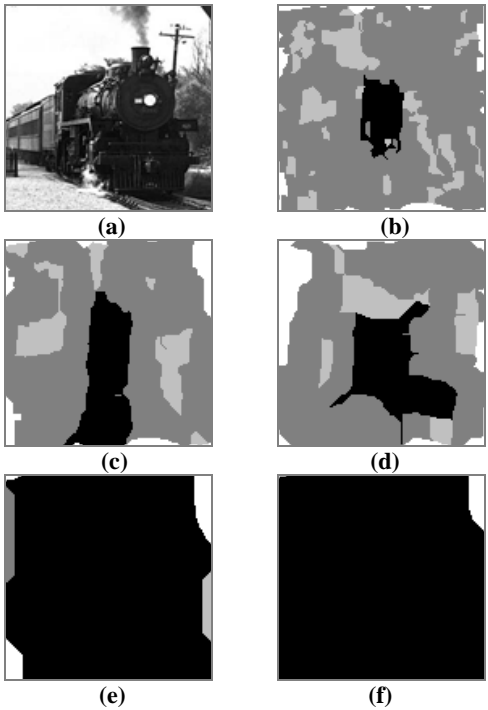

**Fig. 35** The classification maps for the locomotive image with various height distances between the two images in 3D space
(a) the locomotive image
(b) the classification map of total force direction (the distance between the images is 0)
(c) the classification map of total force direction (the distance between the images is 1)
(d) the classification map of total force direction (the distance between the images is 2)
(e) the classification map of total force direction (the distance between the images is 4)
(f) the classification map of total force direction (the distance between the images is 8)

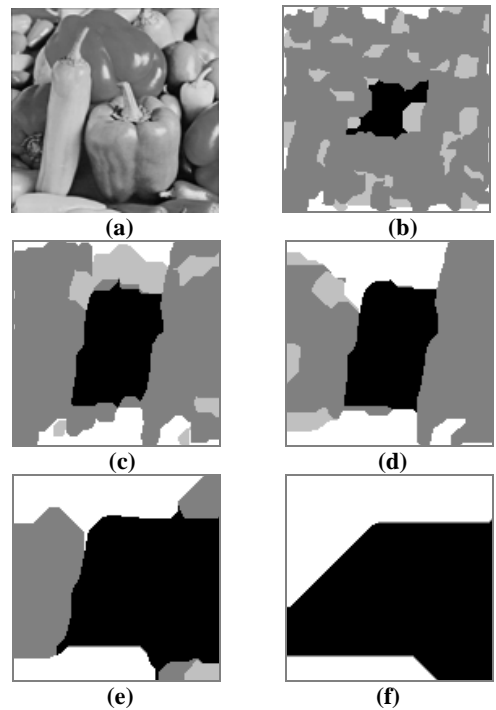

**Fig. 37** The classification maps for the peppers image with various height distances between the two images in 3D space
(a) the peppers image
(b) the classification map of total force direction (the distance between the images is 0)
(c) the classification map of total force direction (the distance between the images is 1)
(d) the classification map of total force direction (the distance between the images is 2)
(e) the classification map of total force direction (the distance between the images is 4)
(f) the classification map of total force direction (the distance between the images is 8)

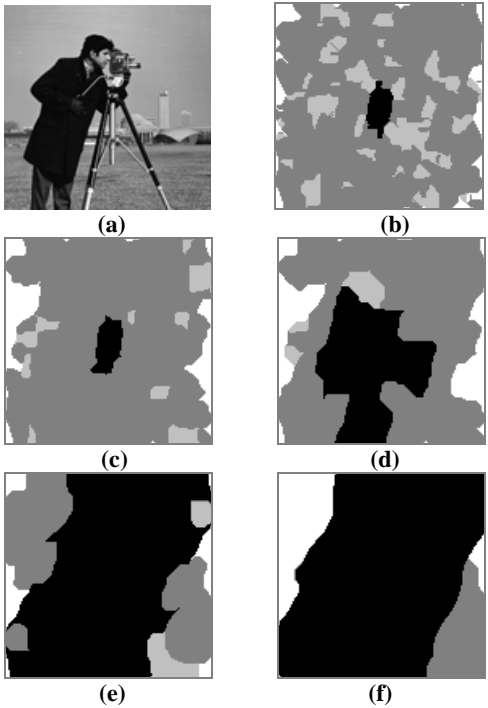

**(a)** **(b)**
**(c)** **(d)**
**(e)** **(f)**

**Fig. 38** The classification maps for the cameraman image with various height distances between the two images in 3D space
(a) the cameraman image
(b) the classification map of total force direction (the distance between the images is 0)
(c) the classification map of total force direction (the distance between the images is 1)
(d) the classification map of total force direction (the distance between the images is 2)
(e) the classification map of total force direction (the distance between the images is 4)
(f) the classification map of total force direction (the distance between the images is 8)

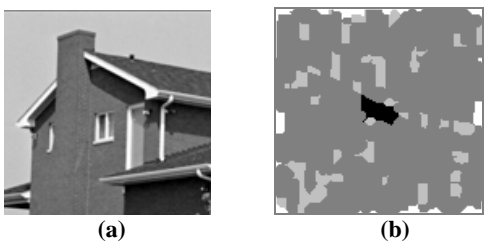

**(a)** **(b)**

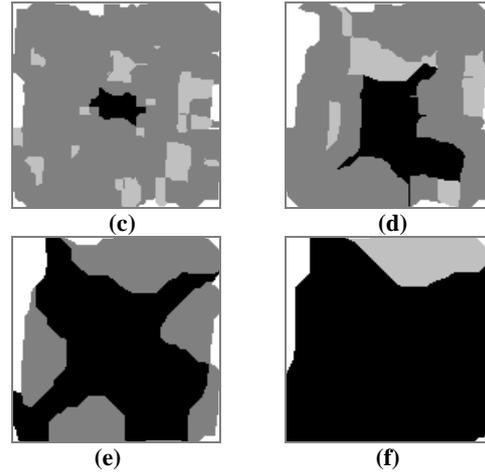

**(c)** **(d)**
**(e)** **(f)**

**Fig. 39** The classification maps for the house image with various height distances between the two images in 3D space
(a) the house image
(b) the classification map of total force direction (the distance between the images is 0)
(c) the classification map of total force direction (the distance between the images is 1)
(d) the classification map of total force direction (the distance between the images is 2)
(e) the classification map of total force direction (the distance between the images is 4)
(f) the classification map of total force direction (the distance between the images is 8)

Based on the above experiment results and analysis, a novel matching method for shift transformation by virtual electromagnetic interaction is summarized as following:

***Step*1** Extract the significant edge lines in the images, and construct the virtual edge current $C_1$ and $C_2$

***Step*2** Calculate the total force on $C_1$ applied by $C_2$

***Step*3** Move the center of image1 one step to the next discrete position according to the discretized direction of the total force

***Step*4** If a balance point is reached, or the moving exceeds the coordinate range of the image, the moving process ends; otherwise, return to ***Step*1**

If the shifted image center is within the convergence point area, the method can successfully achieve image matching for shift transformation.

## 4 Conclusion and Discussion

Magnetic field is a basic and important physical phenomenon. Its unique characteristics may be exploited in the design of novel nature-inspired methods for image processing. Current is a basic source of magnetic field. A macroscopic current in the wire is based on the microscopic movement of electrons. The virtual current proposed in the paper is a kind of reasonable imitation of physical current in the digital images. It is a discrete simulation of the physical current in the discrete image space, which is a discrete flow field running along the edge lines (i.e. the isolines or level curves of grayscale) in the image. In this paper, the virtual edge current is defined based on the significant edge lines which are extracted by a Canny-like operation. The virtual interaction between the significant edge currents is studied by imitating the electro-magnetic interaction between the current-carrying wires, which is then used in image matching for shifting transformation. The preliminary experimental results indicate the effectiveness of the proposed method.

Currently, the approach proposed in this paper does not consider the shift with non-integer displacements. It may be solved by further post-processing on the result of the proposed method. On the other hand, larger shift that makes the center of image1 out of the image coordinate range is not discussed. Further work will study larger range of shift by forming a larger classification map.

Moreover, in practical tasks, the contents of the two images to be matched may not coincide perfectly (i.e. some part in one image does not appear in the other). It is necessary to investigate the effectiveness of the proposed method for different degree of consistency between two images based on the "classification map" of total force, which will be studied in further work. The effect of scaling or rotating between two images will also be investigated in future. On the other hand, the images processed in this paper are of grayscale. For color images, the proposed method can be extended if a reasonable definition of "color gradient" is given, which is also a research topic in future.